\documentclass[letterpaper, 10 pt, conference]{ieeeconf}  

\IEEEoverridecommandlockouts                              

\usepackage{cite}
\usepackage{amsmath,amssymb,amsfonts}
\usepackage{graphicx}
\usepackage{textcomp}
\usepackage{xcolor}
\usepackage[T1]{fontenc}
\usepackage[utf8]{inputenc}
\usepackage[hidelinks]{hyperref}
\usepackage{booktabs}
\usepackage{adjustbox}
\usepackage{url}
\usepackage{color}
\usepackage{soul}

\usepackage{stfloats}
\def\BibTeX{{\rm B\kern-.05em{\sc i\kern-.025em b}\kern-.08em
    T\kern-.1667em\lower.7ex\hbox{E}\kern-.125emX}}

\newcommand{\blind}[2]{\ifblind #1\else #2\fi}

\newif\ifblind
\blindfalse 

\begin{document}

\title{\LARGE \bf Counting What Matters: Reducing Redundant Object Detections in Overlapping UAV Mapping Imagery for Disaster Response}

\author{Max Schulte$^{1}$  Hartmut Surmann$^{1}$ Julien Meine$^{1}$%
\thanks{$^{1}$Westphalian University of Applied Sciences (WH) Gelsenkirchen, Germany 
{\tt\small max.schulte@w-hs.de, hartmut.surmann@w-hs.de, julien.meine@w-hs.de}}%
}

\maketitle
\setcounter{footnote}{1}
\thispagestyle{empty}
\pagestyle{empty}

\begin{abstract}
UAV mapping flights enable rapid aerial reconnaissance of large-scale incident sites in civil protection. However, the image overlap required for orthophoto generation causes every physical object to be detected multiple times across images, while dynamic objects may be omitted from or duplicated in the resulting orthomosaic. Analysis based on the orthophoto alone is therefore structurally insufficient. We address this by detecting objects on the individual images and merging redundant detections across them. This task lies between multi-object tracking and re-identification, as mapping flights form temporally ordered but sparsely sampled sequences with object reappearances across flight strips. We fine-tune YOLO detectors on a new domain-specific nadir dataset from real civil protection operations, roughly doubling mAP50 over the previous system, and propose re-identification approaches based on 3D positions from sparse reconstruction, DINOv3 appearance embeddings, and their combination via constrained greedy clustering. 
On four manually annotated real-world mapping flights, an association-only evaluation using ground-truth detections shows that the combined weighted multi-pass approach provides the best trade-off between association quality and count error.
It is integrated into the open-source web application \blind{\footnote{Repository link omitted for double-blind review.}}{ARGUS\footnote{\url{https://github.com/RoblabWh/argus}}} and visualized as verifiable, correctable object groups for first responders.
\end{abstract}


\section{Introduction}
\label{sec:introduction}

Rapid and reliable reconnaissance is the foundation of tactical decision-making in civil protection \cite{fwdv100}. 
UAVs with RGB or thermal cameras have become an established tool for this assessment and are increasingly deployed by fire brigades and civil protection units \cite{bbk2024}. Beyond live video, autonomous mapping flights, i.e., lawnmower patterns with a nadir-facing camera, provide systematic coverage. The overlapping images are typically fused into orthophotos that serve as up-to-date situation maps.

\begin{figure}
    \centering
    \includegraphics[width=0.98\linewidth]{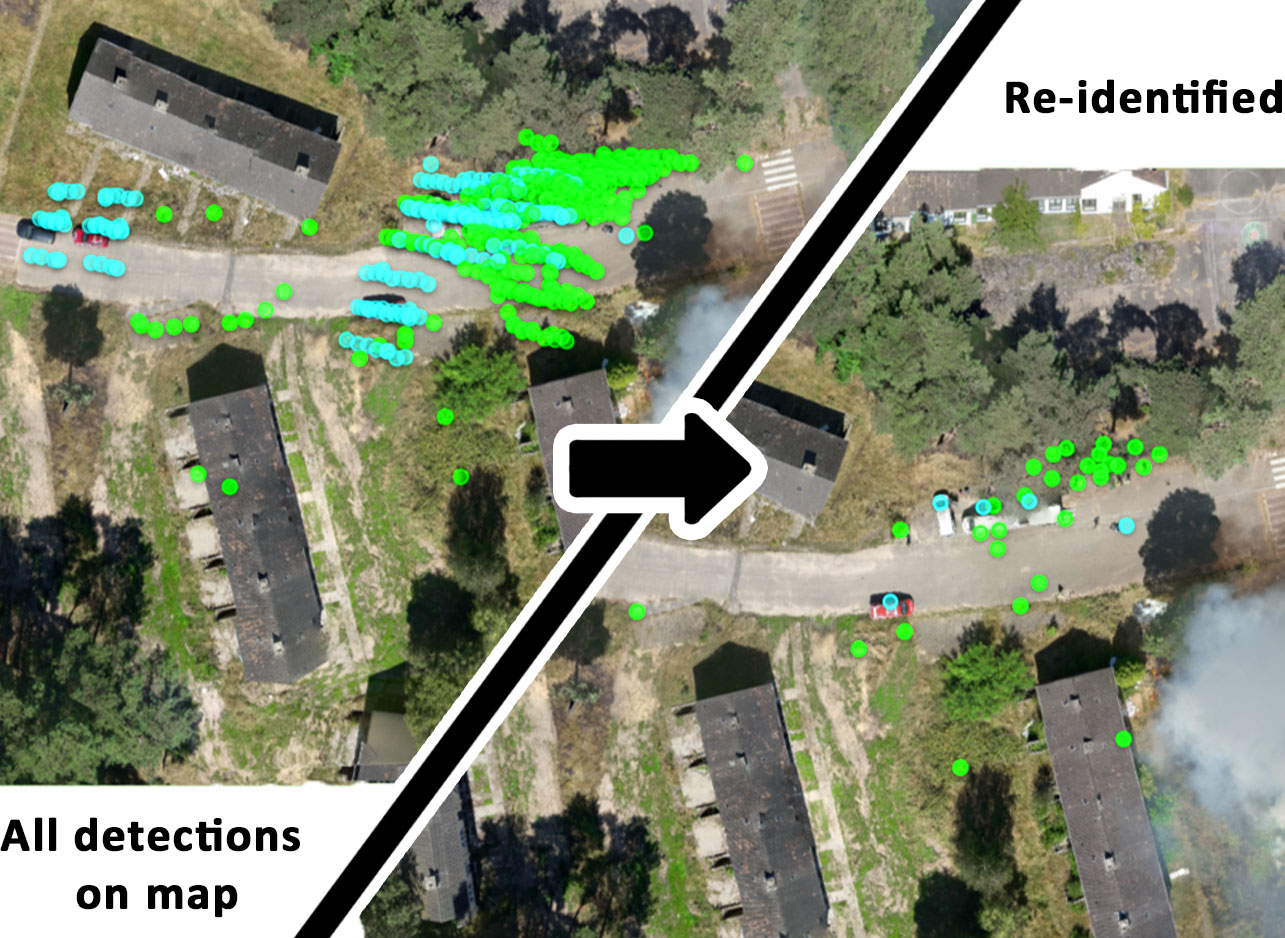}
    \caption{Cluttered map with all roughly geolocated detections (left) and reduced, re-identified objects (right).}
    \label{fig:allToGruped}
\end{figure}

Through regular participation in real deployments, field tests, and exercises with partner fire brigades, we identified a recurring operational need.
Beyond the visual map, responders need a fast answer to seemingly simple questions such as how many people and vehicles are present in the mapped area, and where. Results must be available within minutes on local hardware, must not silently miss people, and must remain verifiable by the operator. Our starting point is\blind{}{ ARGUS \cite{surmann2023},} an open-source web application that manages UAV mapping flights, generates fast orthophotos, and runs object detection on the captured images.

Answering such counting and localization questions automatically is harder than it appears, for two structural reasons. First, detection on the final orthophoto alone is insufficient. Orthomosaic generation implicitly assumes a static scene, so moving objects such as people and vehicles may appear duplicated, incompletely rendered, or be omitted entirely \cite{brack2025}. Occlusion complicates this further, as some objects are visible only from specific viewing angles. Information lost during orthophoto generation cannot be recovered downstream, so objects must be detected on the individual mapping images. Second, per-image detection creates massive redundancy. The 60–80\,\% image overlap required for mapping causes every physical object to be observed in many images. On four manually annotated real-world mapping flights, naively summing per-image detections overestimates the true number of physical objects by a factor of roughly 4 to 20 (depending on the dataset), with each object visible in nine images on average. A useful situational picture therefore requires merging these redundant detections so that each physical object appears on the map only once (Fig.~\ref{fig:allToGruped}).

This merging task occupies a middle ground between multi-object tracking (MOT) and re-identification (re-ID). The images of a mapping flight form a temporally ordered single-camera sequence, which superficially looks like a tracking problem. However, consecutive images are seconds apart, the viewpoint shifts by several meters between frames, and objects routinely leave the field of view and reappear much later in adjacent flight strips of the lawnmower pattern. Sequential tracking assumptions consequently break down. We show that an off-the-shelf tracker (DeepSORT) in fact fails almost completely on this data. Instead, we cast the task as a constrained clustering of detections into physical objects, combining spatial and visual cues under the constraint that detections from the same image must never be merged. This framing also motivates our evaluation protocol, which complements MOT association metrics with clustering-based measures and the application-relevant count error.

We present a complete pipeline from detection to visualization and make the following contributions:

\begin{itemize}
    \item A fine-tuning study of eleven YOLO-based detector variants, including dedicated small-object architectures, on a new domain-specific nadir dataset from real civil protection deployments and exercises, showing that a plain YOLO11x at 1280\,px input resolution outperforms the specialized architectures and roughly doubles mAP50 over the detectors previously used\blind{}{ in ARGUS}.
    \item Position-based, appearance-based, and combined re-identification approaches for merging redundant detections across overlapping mapping images, built on a same-image-constrained union-find clustering. Positions are obtained by backprojecting detections into a sparse 3D reconstruction, and appearance is encoded by DINOv3 embeddings without any identity-annotated training data. We further report a light evaluation of Depth Anything~3 as a learned alternative to classical reconstruction.
    \item 
    An association-only evaluation on four identity-annotated real-world flights, using shared ground-truth detections to isolate association and ablate position, appearance, fusion, and multi-pass variants.
    \item 
    Integration of the pipeline into the open-source ARGUS application, where merged objects are presented as verifiable and correctable groups rather than opaque final counts, together with the released fine-tuned detector.
\end{itemize}

\section{Related Work}
\label{sec:related}

\subsection{Small Object Detection in UAV Imagery}

Object detection in UAV imagery is closely tied to small object detection (SOD). From typical mapping altitudes, people and vehicles occupy only a small portion of the image, and repeated downsampling in convolutional backbones erodes their already sparse features \cite{nisa2025, ning2024}. Current research is dominated by YOLO modifications. A recurring pattern is an additional high-resolution P2 detection head that attaches an earlier backbone feature map to neck and head, preserving fine spatial detail \cite{wang2025sod, lu2025}. DRM-YOLO \cite{bi2026} combines a P2 head with a redesigned backbone, cross-scale feature fusion, and a dynamic detection head, and reports one of the strongest published results on the VisDrone2019 benchmark \cite{zhu2021}, and is one of the few in this area with a ready-to-use published model architecture. As a model-agnostic alternative, slicing-based inference such as SAHI \cite{akyon2022} processes overlapping image tiles separately at the cost of many additional inference calls per image. This was the strategy of the previous\blind{}{ ARGUS} pipeline. Most published SOD gains, however, are demonstrated on nano or small model variants at the default 640 px input resolution, and whether they transfer to large variants at high input resolutions, as required here, is rarely examined. Our detection study addresses exactly this setting.

\subsection{Merging Detections Across Images}

Multi-object tracking is the established approach for associating detections over an image sequence. Tracking-by-detection methods such as DeepSORT and BoT-SORT combine motion prediction with IoU- and appearance-based assignment \cite{ciaparrone2020, aharon2022} and have been applied directly to UAV videos \cite{fonod2025}. However, they presuppose densely sampled sequences with little inter-frame motion. Kostiv et al.~\cite{kostiv2025} study the low-frame-rate limit for UAV footage and observe vanishing box overlap, frequent identity switches, and redundant track initializations, conditions that mapping flights exacerbate through multi-second capture intervals and the lawnmower pattern, under which objects reappear in non-consecutive images.

A second line of work moves the association from image coordinates into a shared spatial frame. Kiefer et al.~\cite{kiefer2023} geolocate detections via camera metadata and accumulate them in likelihood-based memory maps. Yang et al.~\cite{yang2023} project detections onto an assumed ground plane and associate interrupted tracks by world-space distance. Both rely on accurate metadata and flat-scene approximations suited to open water or flat fields. HomView-MOT \cite{ji2026} instead estimates homographies between views and computes a physically more plausible IoU on a common plane, again presupposing planar scenes. Recent surveys see potential in full 3D environments for tracking \cite{li2025}, but existing UAV work operates almost exclusively on dense video with simplified scene models.

Recent feed-forward reconstruction models predict geometry directly from images, but an aerial benchmark found classical COLMAP \cite{schoenberger2016} metrically superior and reported drift/duplicated structures at larger image counts \cite{wu2025}.
Depth Anything~3 \cite{lin2025} is the most recent representative of this family. Its suitability for large nadir mapping flights had not been examined, which we address in Sec.~\ref{sec:reid:position}.

\subsection{Appearance-Based Re-Identification}

Classical person and vehicle re-ID learns discriminative embeddings from identity-annotated datasets \cite{he2021, sun2024}. Such annotations do not exist for nadir mapping imagery from civil protection contexts. Self-supervised foundation models offer a way out. DINOv2 features \cite{oquab2024} have been used without re-ID-specific fine-tuning to correct track identities \cite{lee2025}, to resolve ambiguous associations alongside spatial cues \cite{he2025dino}, and, combined with depth and camera poses, to track object instances through 3D environments \cite{zhao2024}, with the caveat that generic embeddings do not always separate similar instances reliably \cite{huang2025}. DINOv3 \cite{simeoni2025} further improves dense feature quality, explicitly including high-resolution aerial imagery among its demonstrated use cases.

To our knowledge, no prior work addresses cross-image re-identification of people and vehicles in overlapping still images from UAV mapping flights, where neither dense temporal sampling nor identity-annotated training data are available. We close this gap by combining reconstruction-based 3D positions with foundation-model embeddings under an explicit clustering formulation.

\section{Object Detection}
\label{sec:detection}

The re-identification stage operates on per-image detections, so a robust detector with high recall is a prerequisite. Objects missed in individual images may fragment the later association, and in civil protection use, silently overlooked people are the most critical failure mode. As the detectors previously used in \blind{the open-source application}{ARGUS} proved insufficient (Sec.~\ref{sec:eval}), we conducted a dedicated fine-tuning study.

\subsection{Datasets}
\label{sec:detDatsets}

\begin{figure}
    \centering
    \includegraphics[width=0.49\linewidth]{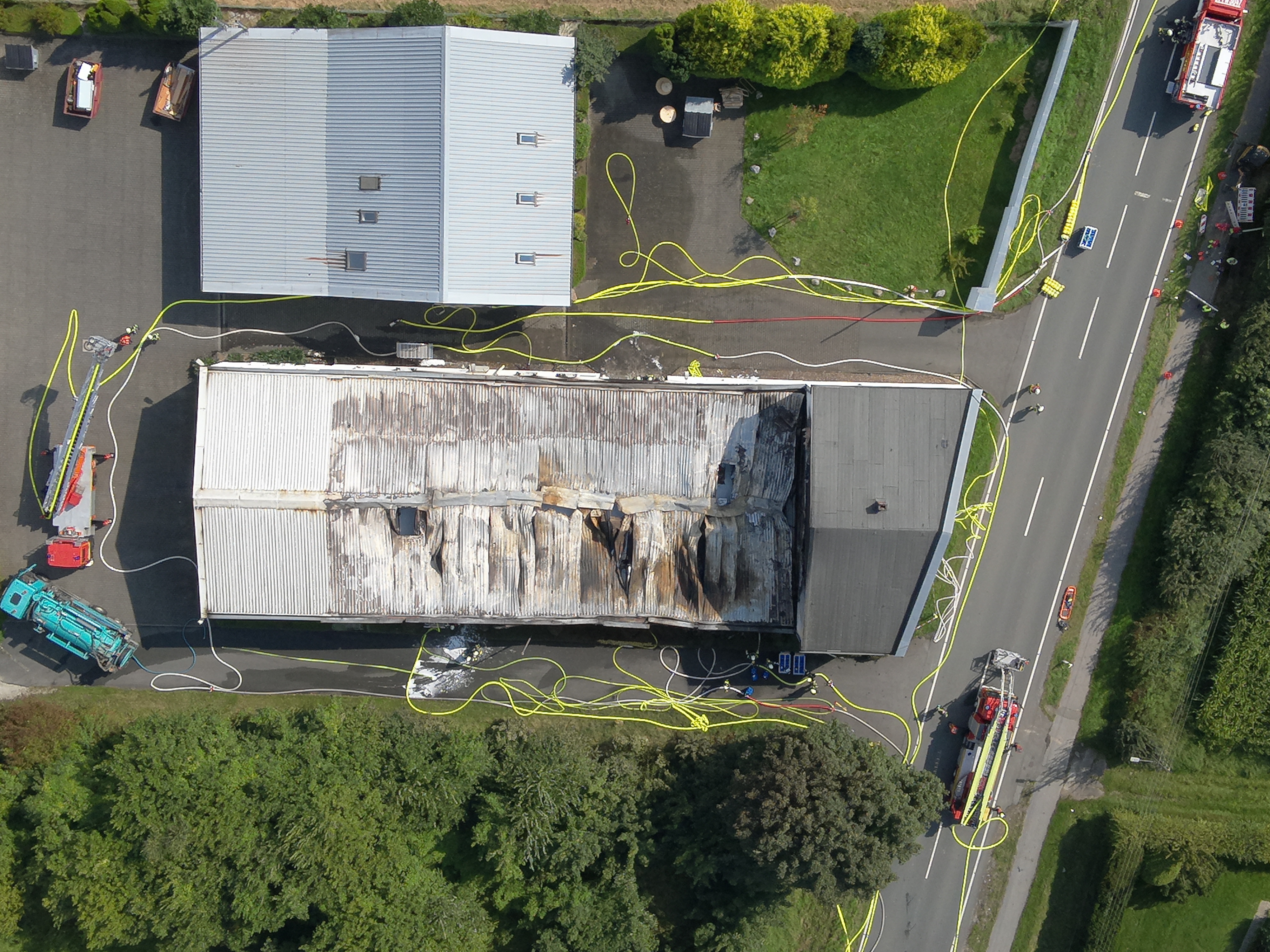}
    \hfill
    \includegraphics[width=0.49\linewidth]{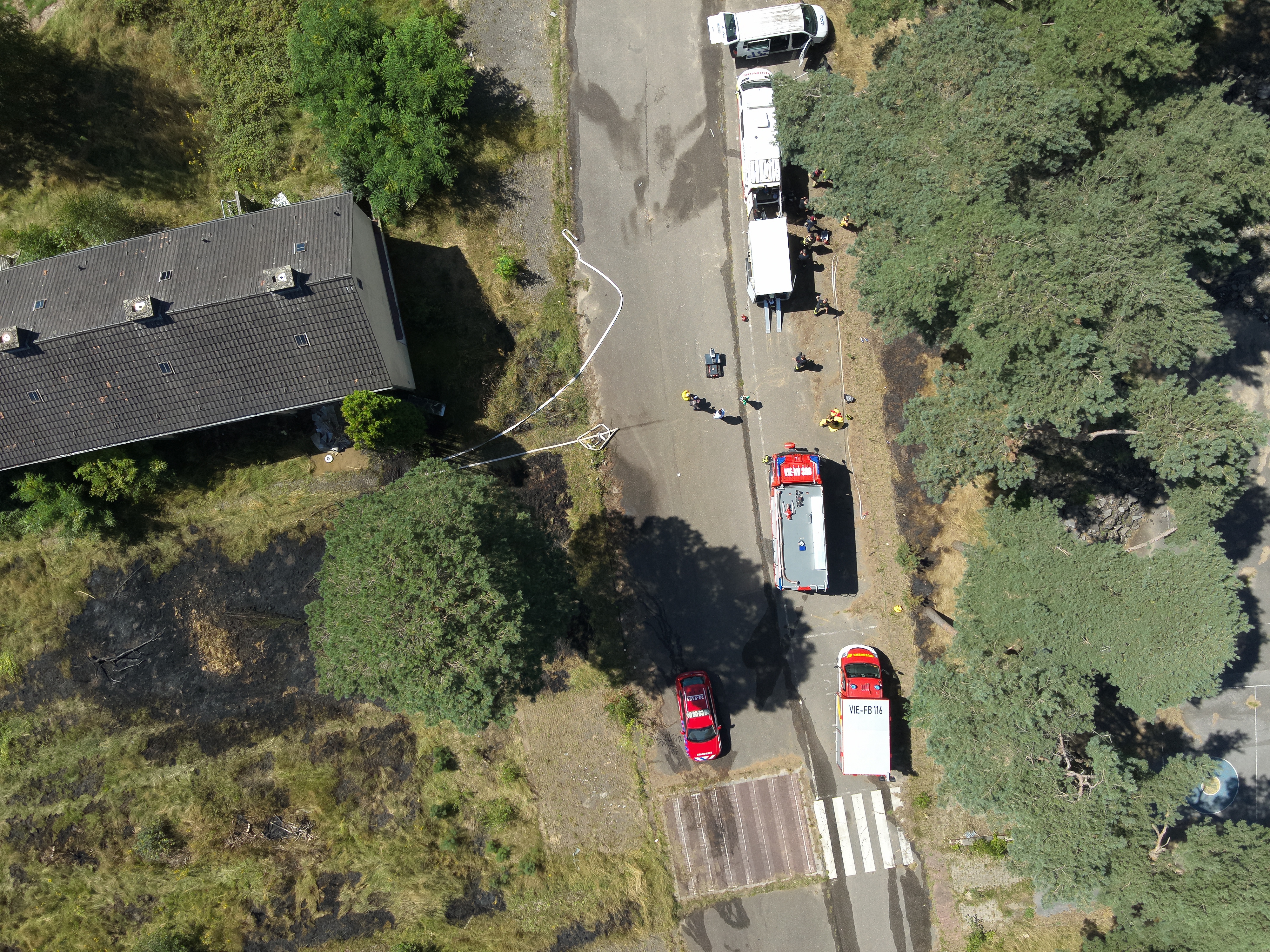} \\[3pt]
    \includegraphics[width=0.49\linewidth]{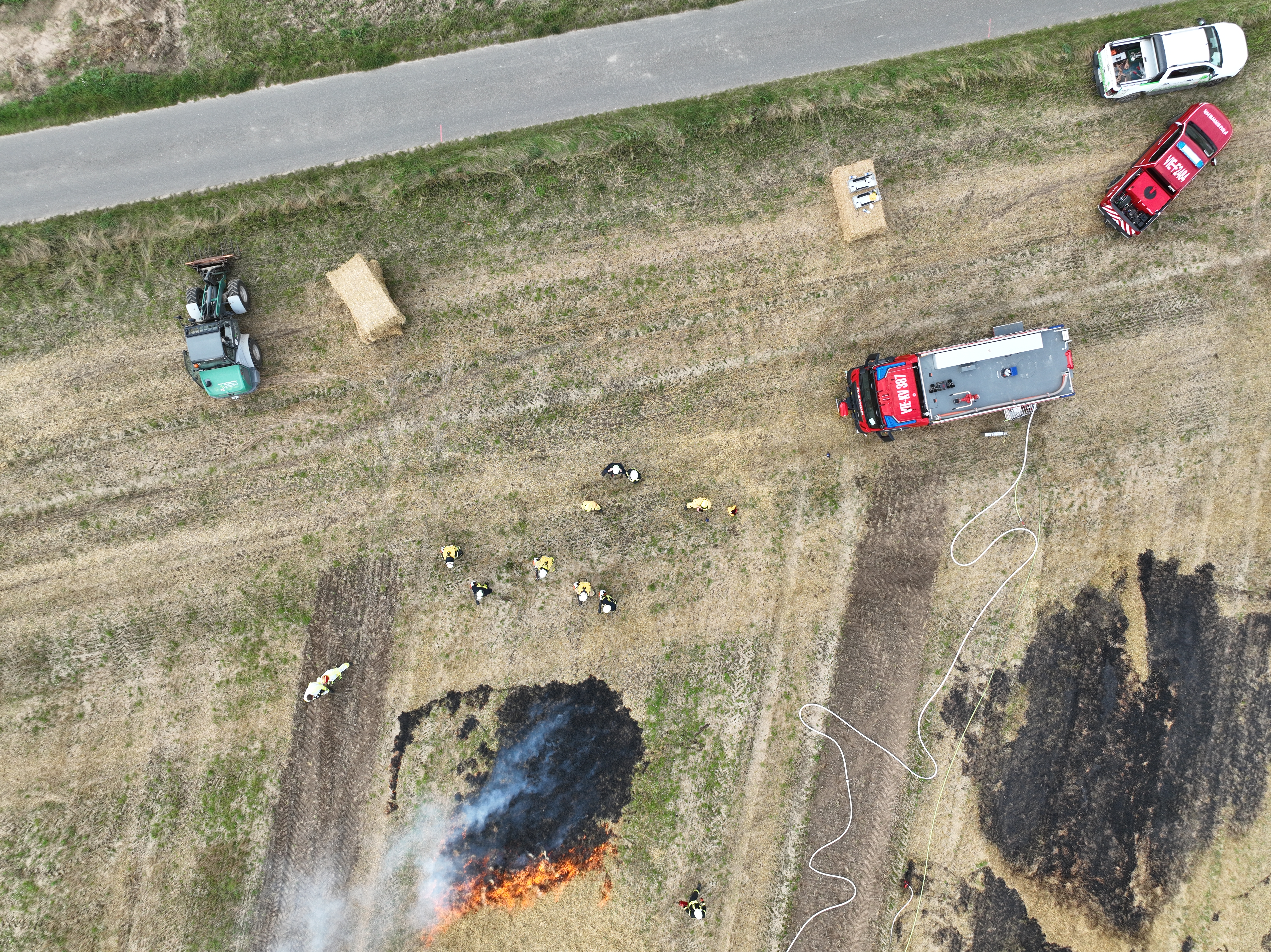} 
    \hfill
    \includegraphics[width=0.49\linewidth]{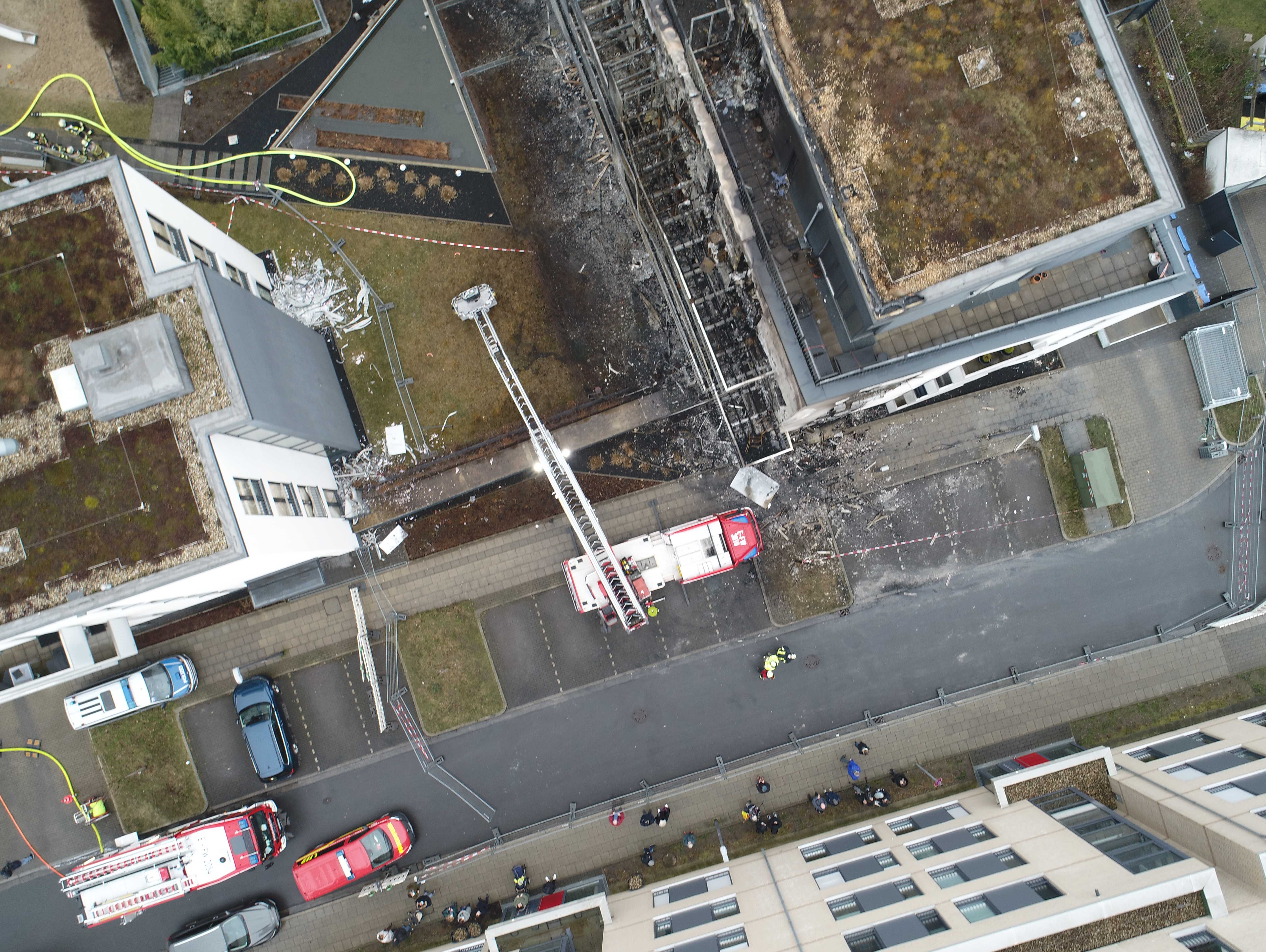}
    \caption{Samples from the ARGUS detection dataset.}
    \label{fig:argusDatasetDetSamples}
\end{figure}

Training combines VisDrone2019 \cite{zhu2021}, the established UAV detection benchmark, with a new domain-specific dataset (Fig. \ref{fig:argusDatasetDetSamples}). Since VisDrone imagery differs from our target domain (predominantly oblique viewing angles, lower resolutions, and general urban scenes), we built \blind{our own}{the ARGUS} dataset. It contains 423 high-resolution nadir images (typically above $4000 \times 3000$\,px) with over 7,200 bounding-box annotations for the classes human and vehicle, collected during real civil protection deployments, field tests, and exercises and complemented by imagery from the drone unit of \blind{a partner fire brigade}{the Bielefeld fire brigade} recorded during real incidents. Images were selected for high variability in scenery, flight altitude, lighting, and object density. Annotation followed a human-in-the-loop process in which an iteratively fine-tuned model proposed boxes that were manually corrected. 
The final domain-specific dataset contains 423 images and is split into 323 training images and a 100-image evaluation split. The split was manually checked for a balanced distribution of scenarios, perspectives, and object densities. Because the number of distinct operational scenarios is limited and the same evaluation split is used to compare candidate detector configurations, the detector results are interpreted primarily as a controlled relative comparison between models rather than as a broad estimate of cross-scenario generalization.

\subsection{Candidate Models and Training}

We compared eleven YOLO-based candidates in three groups: (i) unmodified YOLO11 and YOLO26 in their large and extra-large variants \cite{jocher2023, sapkota2026}, chosen over smaller variants since \blind{the open-source application}{ARGUS} already targets a GPU-equipped local server rather than edge devices; (ii) P2 head extensions of all four, following the prevailing SOD strategy; and (iii) three DRM-YOLO variants of increasing completeness, reproduced from the published model architecture definitions \cite{bi2026} and scaled to the large configuration. 

All candidates were trained in two stages. Pretraining on VisDrone2019, where higher input resolutions consistently improved SOD performance, was followed by fine-tuning on \blind{our own}{the ARGUS} dataset at 1280\,px, matching the later inference resolution, with the detection head adapted to the two target classes. Depending on the available initial weights, candidates were fine-tuned from COCO weights, trained with a temporarily frozen backbone (P2 variants), or trained from scratch (DRM variants). Standard geometric, color, and mosaic augmentations were applied throughout. Since YOLO models are fully convolutional, detection runs natively on the (downscaled) full images, eliminating the recursive tiling of the previous \blind{open-source application}{ARGUS} pipeline together with its many extra inference calls and the merging of detections cut at tile borders.

\section{Re-Identification of Redundant Detections}
\label{sec:reid}

Given per-image detections, the goal is to partition all detections of a mapping flight into groups such that each group contains exactly the observations of one physical object. The investigated approaches differ in the cue used to compare two detections (estimated 3D position, visual appearance, or both) but share a common association mechanism.

\subsection{Constrained Greedy Association}
\label{sec:reid:association}

Unlike classical trackers, we do not process the images as a stream. All images, detections, and metadata of a flight are available jointly, and objects may reappear in temporally distant images, so the association is performed globally over the whole flight. Candidate pairs are generated per object class by the respective cue (e.g., all pairs within a spatial search radius), sorted by descending plausibility, and merged greedily.

Merging is implemented with a union-find structure extended by a hard \emph{same-image constraint} (two detections originating from the same source image, observed simultaneously as distinct bounding boxes, can never depict the same physical object). To enforce this cannot-link condition across transitive merges, the set of contributing image IDs is maintained for every cluster, and a union of two clusters is permitted only if their image-ID sets are disjoint. Otherwise the merge is rejected and the next candidate pair is processed. This keeps the constraint valid not only for individual pairs but across arbitrarily many successive merges, at negligible computational cost. Detections without any accepted link remain singleton clusters and are counted as individual physical objects. The constraint is essential here, since two different objects in the same image could be closer in space or appearance than two observations of the same object in different images. 
Off-the-shelf clustering does not directly enforce the cannot-link constraint and $k$-means additionally requires the number of clusters to be known in advance.

The same-image constraint is treated as a hard physical feasibility condition rather than as a tunable cue: two distinct boxes observed in the same source image cannot correspond to the same physical instance. Its contribution is therefore not reported as a separate score in the ablation below.

\subsection{Position-Based Re-Identification}
\label{sec:reid:position}

Position-based association requires a common spatial reference frame. The coarse georeferencing already available in \blind{the open-source application}{ARGUS}, a bilinear interpolation within the image corners projected onto a flat ground plane using GPS and gimbal metadata, proved too inaccurate. Repeated observations of the same static object scatter by several meters due to GPS latency and 
terrain variation not modeled by the flat-ground assumption. We therefore investigated full 3D reconstructions.

\begin{figure}[b]
    \centering
    \includegraphics[width=1\linewidth]{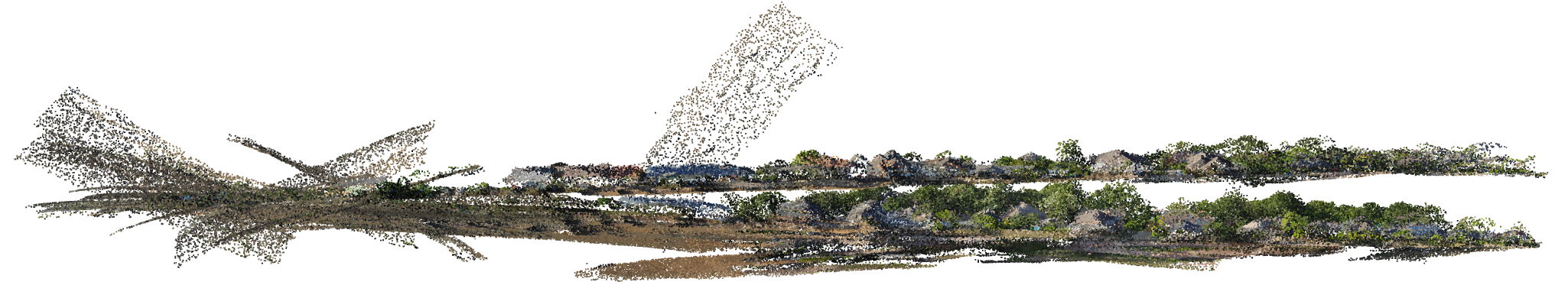}
    \caption{Side view of the DA3 chunked streaming reconstruction, with mispositioned individual images forming intersecting planes (left) and duplicated terrain (top and bottom right).}
    \label{fig:da3sample}
\end{figure}

\subsubsection{Learned Reconstruction with Depth Anything 3}

Motivated by the prospect of faster reconstruction, we evaluated Depth Anything 3 (DA3)~\cite{lin2025} as a learned alternative to classical structure from motion. 
On the test system (Ubuntu 22.04.5, NVIDIA RTX 5080, 16\,GB) only the smaller DA3 variants could process the largest 140-image flight jointly. For the larger flights, both single-pass and official chunked processing produced inconsistent camera geometry and duplicated scene structure after alignment, whereas only the smaller 43-image flight was visually consistent. Because such inconsistencies directly split repeated observations in the spatial reference frame, the geometric re-ID experiments use a classical COLMAP~\cite{schoenberger2016} sparse reconstruction instead. On the same evaluation system, sparse reconstruction required 108--315 s across the four flights.

\subsubsection{Backprojection of Detections}
\begin{figure}[t]
    \centering
    \includegraphics[width=1\linewidth]{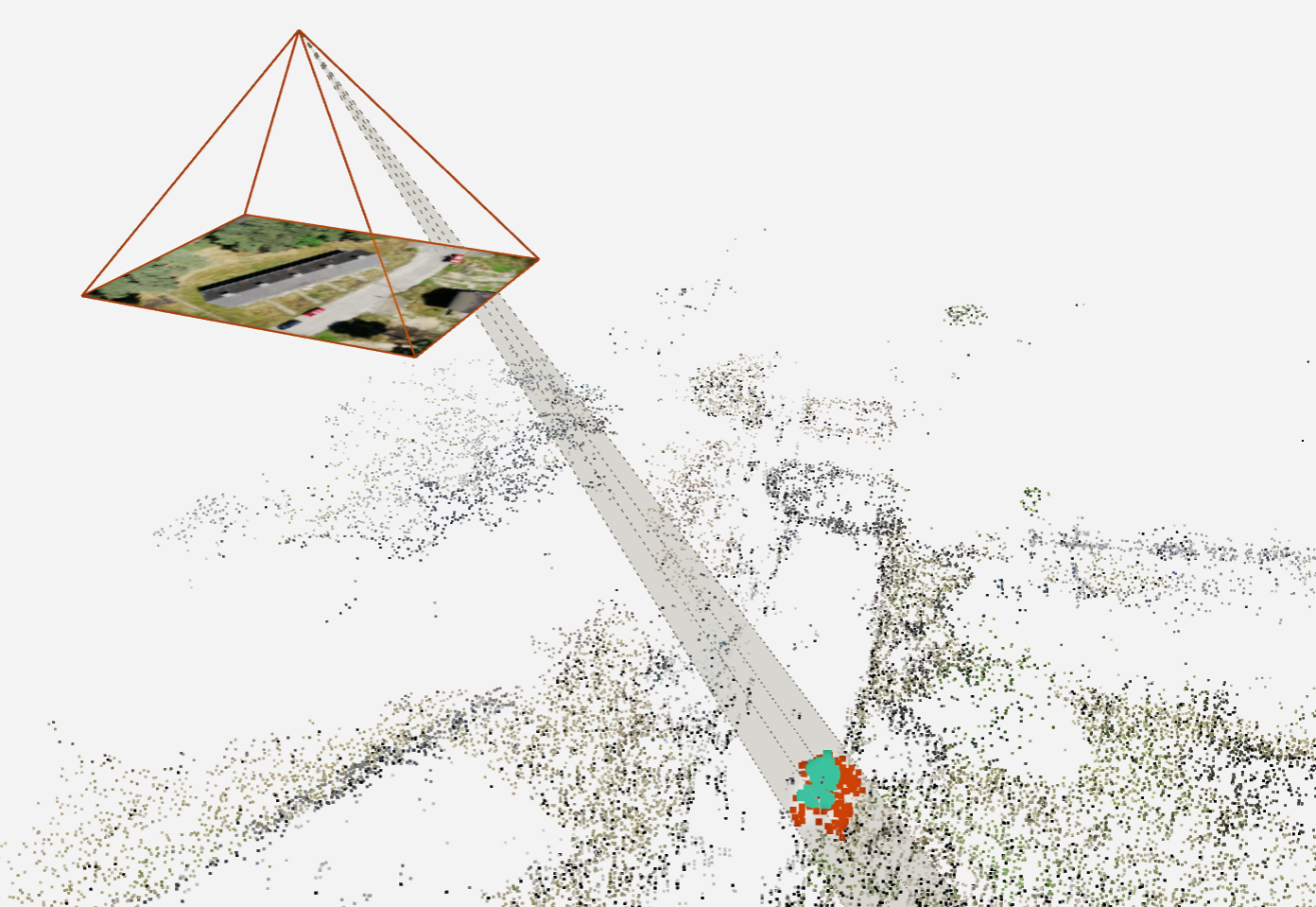}
    \caption{3D backprojection of a detection bounding box into the sparse point cloud.}
    \label{fig:backprojection}
\end{figure}
For each detection, a 3D position is estimated from its bounding box, the reconstructed camera pose, and the sparse point cloud (Fig.~\ref{fig:backprojection}). Since a single ray through the box center rarely hits a sparse 3D point, the primary method casts rays through the four box corners, forming a viewing frustum, and collects all sparse points inside it. After outlier filtering, the points closest to the camera, i.e., those most likely belonging to the visible object surface rather than the background, are averaged to obtain a robust position estimate. If the frustum contains insufficient points, we successively enlarge it, intersect the center ray with a locally fitted RANSAC ground plane, and finally fall back to a flat-ground projection using the refined COLMAP pose and relative flight altitude. 
The backprojection thus remains anchored to the actual reconstruction as long as possible and degrades gracefully in feature-poor regions.


\subsubsection{Clustering of 3D Positions}
\begin{figure}
    \centering
    \includegraphics[width=1\linewidth]{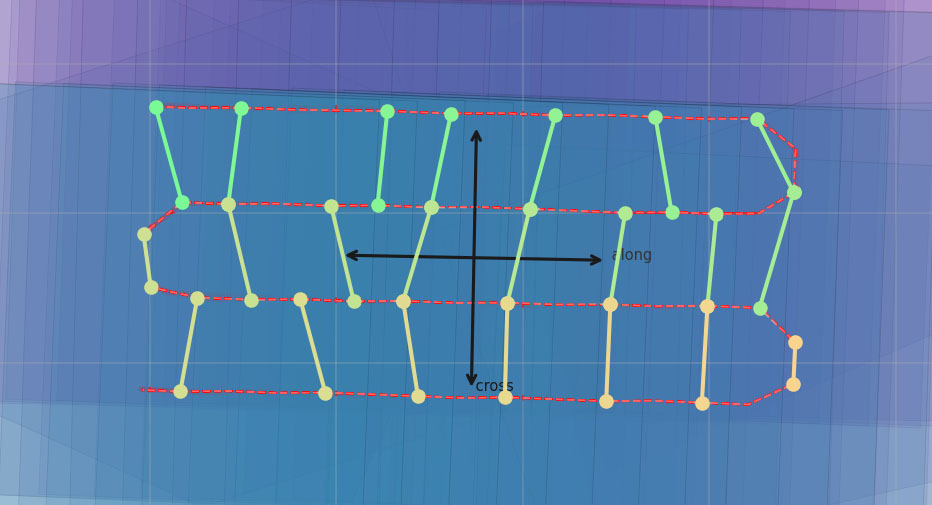}
    \caption{Bridges connecting flight pattern lines}
    \label{fig:bridges}
\end{figure}

Three purely geometric variants were implemented.
The \emph{global} variant builds candidate pairs within class-specific search radii of 4.5\,m for people and 10\,m for vehicles. These comparatively permissive radii were selected to tolerate both object motion and 3D backprojection error, including displaced position estimates for partially visible objects near image borders; smaller radii increase fragmentation, whereas larger radii increase the risk of false merges. Candidate pairs are then processed in order of increasing 3D distance using the constrained union-find procedure from \ref{sec:reid:association}.
It corresponds to greedy single-linkage clustering with cannot-link constraints. The \emph{sequential} variant follows the tracking-by-detection paradigm and associates only detections of consecutive images. The \emph{bridged} variant adds a second pass exploiting the known lawnmower pattern. Image pairs with strong lateral overlap across neighboring flight strips are identified as \emph{bridge pairs} (Fig.~\ref{fig:bridges}) and associated as well, reconnecting tracklets that the flight pattern had split in the \emph{sequential} run.


\subsubsection{Homography-Based Variant}

As a reconstruction-free alternative inspired by~\cite{ji2026}, we estimate planar homographies between consecutive image pairs from ORB feature matches with RANSAC, masking out detected objects so that dynamic content does not alter the background motion estimate. Bounding boxes are projected into the paired image and associated greedily by IoU (threshold 0.3), sequentially and optionally with bridge pairs. This avoids any reconstruction but inherits the flat-scene assumption of the homography, which becomes problematic for scenes with pronounced height structure.

\subsection{Appearance-Based Re-Identification}
\label{sec:reid:appearance}
As identity-annotated re-ID training data for nadir mapping imagery does not exist, the appearance-based approach relies on a general-purpose visual foundation model used purely for inference. Each detection is cropped from the original image with a small context margin and encoded by DINOv3~\cite{simeoni2025} (ViT-B/16). The pooled CLS token yields a 768-dimensional, L2-normalized embedding, so similarity reduces to the cosine (dot product). To exclude implausible matches and limit computation, candidate pairs are prefiltered by the coarse\blind{}{ ARGUS} ground positions with a more generous radius. 
Pairs with cosine similarity of at least 0.65 are merged greedily in descending similarity. This threshold was selected empirically and is slightly stricter than the value of 0.6 reported in~\cite{zhao2024} for DINOv2-based instance association; an independent sensitivity study was not performed.

\subsection{Combined Position and Appearance}
\label{sec:reid:combined}

Position and appearance have complementary failure modes. Geometry alone confuses nearby or moving objects, whereas appearance alone confuses similar-looking ones. A simple combined variant keeps the 3D distance as the primary criterion and uses visual similarity only as a sanity check that vetoes geometrically close but visually dissimilar links. The best-performing variant fuses both cues into a weighted distance $d_\mathrm{w} = d_\mathrm{3D} \cdot (1.6 - \mathrm{sim})$, which contracts visually similar pairs and repels dissimilar ones, and applies it in a progressive multi-pass scheme. 
Candidate links are processed over several passes with successively relaxed appearance thresholds (0.95 down to 0.61) and growing spatial search radii. The intent is to commit highly confident, mostly static associations first and admit less certain links, including moderately displaced moving objects, only after the cluster structure has stabilized. A final cleanup pass revisits small residual clusters and attempts to merge them with visually compatible clusters in an enlarged search region while preserving the same-image constraint. Motion is therefore accommodated heuristically but is not explicitly modeled; the method does not use inter-image time differences or a motion model.

\section{Evaluation}
\label{sec:eval}

\subsection{Object Detection}
\label{sec:eval:detection}

All candidates were evaluated on the held-out\blind{}{ ARGUS} test split at 1280\,px input resolution, with mAP50 as primary metric and particular attention to recall, since missed people are the most critical failure mode. Table~\ref{tab:detection} summarizes the leading variants of each architecture group alongside the detectors previously used in \blind{the open-source application}{ARGUS}.

\begin{table}
\caption{Detection results on the evaluation split (1280\,px input). Top: YOLO baseline and P2 variants alongside the best DRM variant (pretrained on VisDrone2019 at 960\,px). Bottom: detectors previously used in ARGUS at their native inference resolutions. Inference time at batch size 1.}
\label{tab:detection}
\centering
\small
\begin{adjustbox}{max width=\linewidth}
\begin{tabular}{lccccc}
    \toprule
    Model & P & R & mAP50 & mAP50:95 & $t$\,[ms] \\
    \midrule
    YOLO11l            & 0.828 & 0.812 & \textbf{0.869} & 0.605 & \textbf{12.8} \\
    YOLO11l-P2         & 0.752 & 0.746 & 0.787 & 0.546 & 17.0 \\
    YOLO11x            & 0.835 & \textbf{0.820} & 0.864 & \textbf{0.617} & 23.1 \\
    YOLO11x-P2         & 0.762 & 0.729 & 0.778 & 0.534 & 32.2 \\
    YOLO26l            & 0.852 & 0.776 & 0.849 & 0.602 & \textbf{12.7} \\
    YOLO26l-P2         & 0.753 & 0.642 & 0.715 & 0.494 & 17.1 \\
    YOLO26x            & \textbf{0.875} & 0.777 & 0.868 & 0.610 & 23.5 \\
    YOLO26x-P2         & 0.763 & 0.694 & 0.743 & 0.518 & 32.9 \\
    DRM-MNxt-Dymic-l   & 0.799 & 0.730 & 0.781 & 0.541 & 67.4 \\
    \midrule
    Deformable-DETR (prev.) & 0.594 & 0.429 & 0.449 & 0.191 & 78.8 \\
    YOLOS-small (prev.)     & 0.377 & 0.288 & 0.296 & 0.114 & 49.0 \\
    \bottomrule
\end{tabular}
\end{adjustbox}
\end{table}

Two findings stand out. First, the dedicated SOD architectures provide no benefit in this setting. All tested P2 variants score below their unmodified counterparts (e.g., YOLO11x drops from 0.864 to 0.778 mAP50) while requiring up to ${\sim}73\,\%$ more VRAM and ${\sim}40\,\%$ longer inference. The best DRM variants reach only 0.781--0.782 mAP50 at substantially higher resource cost (1{,}545\,MB and 67\,ms versus 466\,MB and 12.8\,ms for YOLO11l). A plausible explanation is that at 1280\,px input, people and vehicles are already sufficiently represented on the regular feature maps of the large model variants, so an additional high-resolution head mainly adds complexity and newly initialized parameters. SOD-specific extensions, typically demonstrated on nano and small models at 640\,px, may thus not transfer to this setting. Second, class-wise analysis shows vehicles are detected far more reliably (mAP50 0.96--0.97) than people (0.75--0.78, recall 0.65--0.70), which are smaller, vary more in pose, and are more easily occluded. Person detection remains the bottleneck.

Among the leading models, the fine-tuned YOLO11x was selected for \blind{the software}{ARGUS}. It achieves the highest overall recall, the highest person recall, and the best mAP50:95, and its additional inference cost over YOLO11l amounts to only seconds for a typical flight on the targeted local server hardware. Compared to the strongest previous \blind{}{ ARGUS} detector, YOLO11x roughly doubles mAP50 (0.449 to 0.864) at less than a third of the inference time, while additionally eliminating the recursive tiling and its many extra inference calls per image.

\subsection{Re-Identification}
\label{sec:eval:reid}

\begin{table}[!t]
\caption{Re-ID Evaluation flights: images, per-image GT annotations, unique physical objects, and share of unique objects among annotations.}
\label{tab:datasets}
\centering
\small
\begin{tabular}{lcccc}
    \toprule
    ID & Images & Annot. & Objects & Share [\%] \\
    \midrule
    D1 & 43  & 241 & 58 & 24.1 \\
    D2 & 140 & 802 & 93 & 11.6 \\
    D3 & 78  & 405 & 37 & 9.1 \\
    D4 & 67  & 507 & 28 & 5.5 \\
    \bottomrule
\end{tabular}
\end{table}

\subsubsection{Datasets and Protocol}

The association approaches are evaluated on four complete mapping flights from real operations and exercises that were manually annotated with cross-image physical-object identities (Table~\ref{tab:datasets}). These flights are distinct in role from the 323/100-image detector training/evaluation split of Sec. \ref{sec:detDatsets}: here, complete flights are required because repeated sightings of the same physical object must be linked across images. \emph{D1} covers an urban industrial site with two elongated flight strips, a loose crowd, and vehicles in street canyons between large buildings. \emph{D2} covers the partially flooded, evacuated residential area and is dominated by static vehicles with comparatively few people. \emph{D3} shows a vegetation-fire exercise with many visually similar responders in close proximity, while \emph{D4} contains many responders between buildings with strong vegetation occlusion and pronounced inhomogeneous motion.

\begin{table*}[!b]
\caption{Association results and component ablation on ground-truth
boxes. PW = pairwise F1, CE = count error. All fusion variants
below the second midrule use global 3D positions. Best value per
metric and dataset is shown in bold.}
\label{tab:reid}
\centering
\small
\begin{adjustbox}{max width=\textwidth}
\begin{tabular}{lccc@{\hskip 5mm}ccc@{\hskip 5mm}ccc@{\hskip 5mm}ccc}
    \toprule
     & \multicolumn{3}{c}{D1}
     & \multicolumn{3}{c}{D2}
     & \multicolumn{3}{c}{D3}
     & \multicolumn{3}{c}{D4} \\
    \cmidrule(r{5mm}){2-4}
    \cmidrule(r{5mm}){5-7}
    \cmidrule(r{5mm}){8-10}
    \cmidrule{11-13}
    Approach
    & PW & AssA & CE\,[\%]
    & PW & AssA & CE\,[\%]
    & PW & AssA & CE\,[\%]
    & PW & AssA & CE\,[\%] \\
    \midrule

    Homography sequential
    & 0.366 & 0.451 & 189.7
    & 0.517 & 0.480 & 154.8
    & 0.357 & 0.321 & 337.8
    & 0.263 & 0.199 & 657.1 \\

    Homography seq.\,+\,bridges
    & 0.366 & 0.451 & 189.7
    & 0.888 & 0.820 & 77.4
    & 0.694 & 0.616 & 227.0
    & 0.598 & 0.406 & 567.9 \\

    \midrule

    3D sequential
    & 0.751 & 0.745 & 43.1
    & 0.522 & 0.484 & 152.7
    & 0.456 & 0.397 & 132.4
    & 0.473 & 0.360 & 260.7 \\

    3D sequential\,+\,bridges
    & \textbf{0.932} & \textbf{0.904} & 17.2
    & 0.869 & 0.799 & 69.9
    & 0.716 & 0.664 & 37.8
    & 0.796 & 0.638 & 153.6 \\

    3D global
    & 0.896 & 0.872 & \textbf{1.7}
    & 0.904 & 0.870 & \textbf{18.3}
    & 0.823 & 0.748 & 24.3
    & 0.794 & 0.658 & \textbf{42.9} \\

    DINOv3
    & 0.874 & 0.818 & 43.1
    & 0.716 & 0.662 & 96.8
    & 0.789 & 0.703 & 43.2
    & 0.773 & 0.620 & 167.9 \\

    \midrule

    3D global\,+\,DINOv3
    & 0.841 & 0.789 & 41.4
    & 0.840 & 0.781 & 80.6
    & 0.849 & 0.761 & 48.6
    & 0.799 & 0.662 & 150.0 \\

    3D global\,+\,DINOv3 multi-pass
    & 0.897 & 0.858 & 34.5
    & 0.876 & 0.818 & 67.7
    & \textbf{0.893} & \textbf{0.834} & 32.4
    & 0.810 & 0.658 & 135.7 \\

    3D global\,+\,DINOv3 weighted
    & 0.913 & 0.879 & 19.0
    & 0.904 & 0.856 & 39.8
    & 0.787 & 0.707 & 35.1
    & 0.814 & 0.667 & 103.6 \\

    3D global\,+\,DINOv3 weighted multi-pass
    & 0.925 & 0.882 & 10.3
    & \textbf{0.920} & \textbf{0.881} & 29.0
    & 0.868 & 0.800 & \textbf{16.2}
    & \textbf{0.860} & \textbf{0.737} & \textbf{42.9} \\

    \bottomrule
\end{tabular}
\end{adjustbox}
\end{table*}

To isolate association quality from detection errors, every method receives the same manually corrected ground-truth bounding boxes. Consequently, the reported re-ID results quantify association quality rather than end-to-end detection-and-counting performance. 
This results in a $\mathrm{DetA}=1$, so HOTA reduces to $\sqrt{\mathrm{AssA}}$. We therefore report AssA directly.
We additionally report pairwise F1 and count error, defined as
\(CE = \frac{|K_{\mathrm{pred}}-K_{\mathrm{gt}}|}{K_{\mathrm{gt}}}\cdot100\,\%\), where \(K_{\mathrm{pred}}\) and \(K_{\mathrm{gt}}\) denote the predicted and ground-truth numbers of physical-object clusters, respectively.
CE is an application-relevant summary of the final number of groups, but can hide compensating merge and fragmentation errors and is therefore interpreted jointly with association metrics.

\subsubsection{Baselines}

The unmodified DeepSORT baseline fails almost completely under the multi-second capture gaps and large inter-frame displacement of the mapping flights, producing 1,881 identities for 216 physical objects in aggregate (CE approximately 770\,\%). Mosaic-footprint clipping based on the ARGUS orthomosaic can yield deceptively small count errors on D1--D3 (10--14\,\%) but performs no identity assignment and may omit objects entirely. In D4, only 7 of 28 physical objects remain represented after clipping. These results motivate explicit cross-image association and also show why CE must not be interpreted in isolation.

\subsubsection{Position, Appearance, and Combination}

Homography-based association remains fragile in non-planar scenes, with CE between 77\,\% and 568\,\% even when bridge pairs are added. In one case, the homography over-transforms the bounding boxes because it describes the transformation between the elevated flat roofs rather than the lower street plane. Sequential 3D association also fragments identities, whereas global association in the common 3D frame is substantially more robust. DINOv3 alone achieves competitive association quality without re-ID-specific training but remains strongly fragmented (CE 43–168\,\%).

The ablation in Table \ref{tab:reid} shows that simply combining geometry and appearance does not guarantee improvement: simple fusion and single-pass weighting behave inconsistently. Progressive merging improves stability, and weighting is most effective when combined with the multi-pass scheme. The weighted multi-pass variant achieves the strongest association among the combined methods on three of four flights and improves AssA over global 3D on all four flights, with CE ranging from 10.3\,\% to 42.9\,\%.
We therefore use it in ARGUS, with the DINOv3-based appearance-only association as a reconstruction-free fallback.

\section{Integration and Visualization \blind{}{ in ARGUS}}
\label{sec:integration}

The selected YOLO11x detector and the two recommended re-ID approaches were integrated into the open-source\blind{}{ ARGUS} application. Once the reconstruction is available, the weighted multi-pass combination can be applied, with the appearance-only approach as fallback. Until re-ID results arrive, the map falls back to the mosaic-footprint-clipped detections as a fast provisional view. Detection and sparse reconstruction run as independent asynchronous workers. A completed reconstruction is persisted and reused for later re-ID runs, so the multi-minute reconstruction cost is not incurred each time detections are regrouped.

\begin{figure}[!t]
    \centering
    \includegraphics[width=1\linewidth]{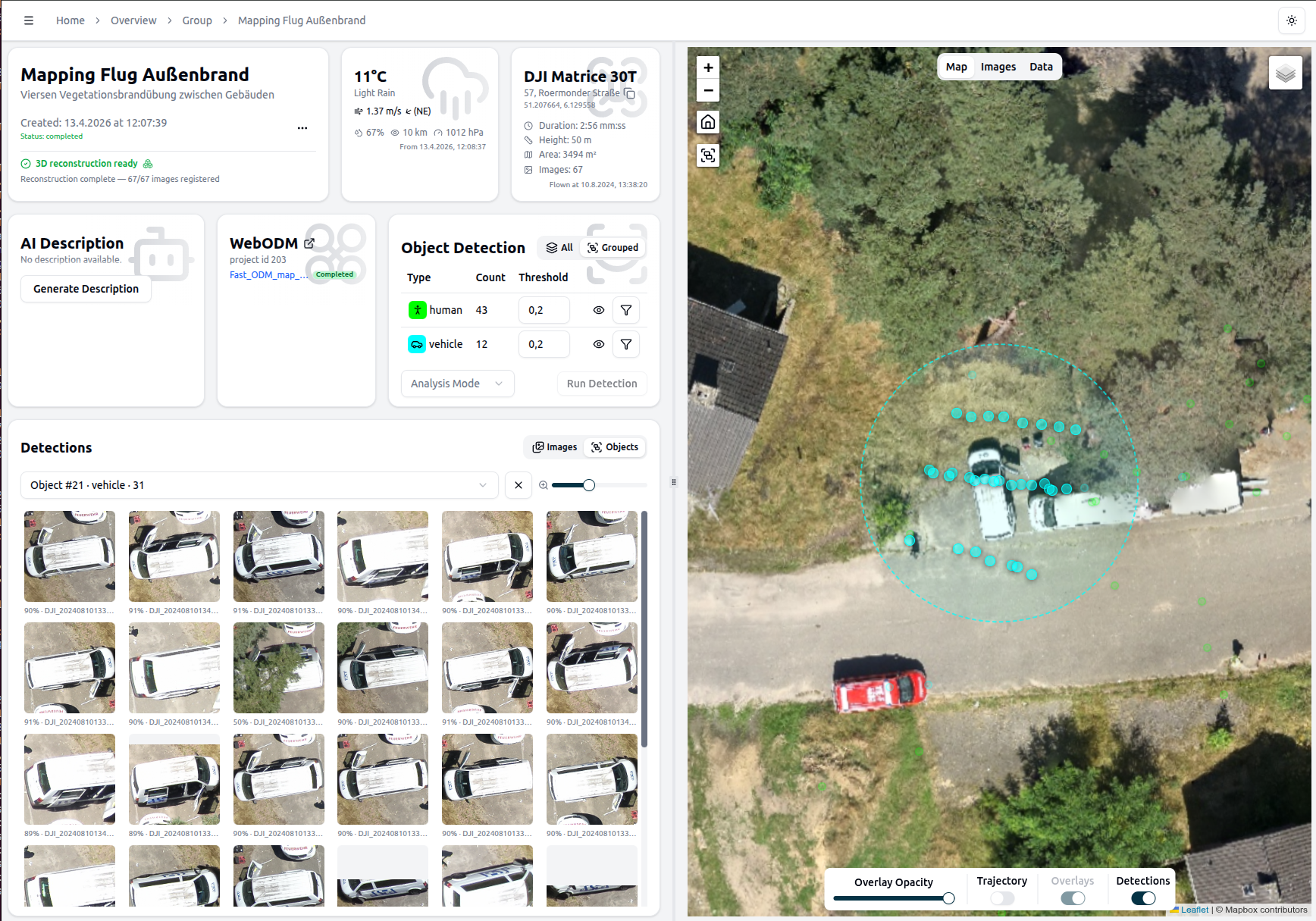}
    \caption{Grouped detections marked on the map with the corresponding cropped-object gallery.}
    \label{fig:argus}
\end{figure}

Since no re-ID approach is error-free, the visualization is designed to be verifiable and correctable rather than opaque. Each physical object appears as a single map marker annotated with the number of merged sightings. Selecting it reveals the individual detections and a gallery of all associated image crops (Fig.~\ref{fig:argus}), so an operator can confirm at a glance whether the group indeed shows one object. Erroneous groups can be split or merged and individual detections reassigned directly on the map, keeping the human in the loop for consequential decisions such as confirming a person count. The resulting count is therefore treated as operator-verifiable decision support rather than as an autonomous tactical decision.

\section{Discussion}
\label{sec:discussion}



Within the evaluated scenarios, domain-specific high-resolution fine-tuning yielded larger gains than the tested SOD modifications, while global constrained clustering was more robust than sequential tracking. The ablation further shows that appearance and geometry are complementary, with weighted multi-pass association providing the most consistent combined results.

The remaining count errors require careful interpretation. We do not regard the observed 10.3--42.9\,\% CE as validated for autonomous operational counting. The re-ID evaluation uses ground-truth boxes and therefore measures association error only. 
Detector false positives can create extra clusters and missed detections can remove or fragment objects, so end-to-end count accuracy remains unquantified. Within association, fragmentation causes overcounting, while false merges can cause undercounting. 


Dynamic objects remain challenging because motion and localization error both appear as spatial displacement.
The present method has no explicit motion model and does not use the capture-time differences between images. Instead, moderate motion is accommodated heuristically through permissive search radii, appearance cues, progressive multi-pass relaxation, and the final cleanup of small residual clusters. 
Motion beyond the search region or strong appearance changes can fragment identities, whereas nearby, visually similar objects may be falsely merged; D4, with pronounced motion and vegetation occlusion, illustrates this limitation.

Generalization remains limited by four re-ID flights and a single 323/100 detector split without cross-validation or leave-one-scenario-out testing. 
The datasets cover qualitatively different environments but are too limited in number to establish broader generalization across flight altitudes, lighting conditions, terrain types, and disaster scenarios.
Association parameters were selected empirically. Stricter settings increase fragmentation while permissive ones risk false merges, but no independent sensitivity study or cleanup ablation was performed. Finally, the combined method depends on successful sparse reconstruction (108–315 s here), although ARGUS caches it and can fall back to appearance-only association.

\section{Conclusion}
\label{sec:conclusion}


We presented a pipeline for turning redundant per-image detections from overlapping UAV mapping flights into an operator-verifiable object overview for civil protection. On the detector evaluation split, domain-specific fine-tuning of standard YOLO11 models at high input resolution roughly doubles mAP50 over the detectors previously used in ARGUS without requiring dedicated SOD architectures. For cross-image association, experiments on ground-truth detections from four complete mapping flights show that global 3D position is substantially more reliable than sequential tracking assumptions, while DINOv3 appearance features provide complementary evidence for nearby, occluded, or moderately moving objects. The component ablation further shows that weighted distance and progressive multi-pass association are most effective when used together. The final variant reaches pairwise F1 values of 0.86--0.93 and association-stage count errors of 10.3--42.9\,\% on the four flights.

These values should not be interpreted as end-to-end operational counting accuracy because detector errors were intentionally excluded from the re-ID evaluation. ARGUS therefore exposes the grouped observations and their image crops for inspection and manual correction instead of treating the predicted count as authoritative. Future work should evaluate the full detector-association pipeline end to end on additional scenarios, quantify parameter sensitivity, exploit inter-image time gaps as motion priors for dynamic objects, and revisit learned reconstruction as models such as Depth Anything~3 mature toward consistency at mapping scale.

\ifblind
\section*{Acknowledgment}
Removed for double-blind review.
\else
\section*{Acknowledgment}
This work is funded by the German Federal Ministry of Research, Technology and Space (BMFTR) under grant 13N16478, project "E-DRZ: Establishment of the German Rescue Robotics Center", https://rettungsrobotik.de.
\fi

\bibliographystyle{IEEEtran}
\bibliography{bibliography}

@misc{Wang2025sod,
      title={SOD-YOLO: Enhancing YOLO-Based Detection of Small Objects in UAV Imagery}, 
      author={Peijun Wang and Jinhua Zhao},
      year={2025},
      eprint={2507.12727},
      archivePrefix={arXiv},
      primaryClass={cs.CV},
      url={https://arxiv.org/abs/2507.12727}, 
}

@misc{fwdv100,
  title        = {{Feuerwehr-Dienstvorschrift 100: Führung und Leitung im Einsatz -- Führungssystem}},
  year         = {1999},
  author = {AFKzV},
  note         = {{Ausschuss für Feuerwehrangelegenheiten, Katastrophenschutz und zivile Verteidigung (AFKzV)}. Stand: März 1999, mit Änderungen aus 2005},
  howpublished = {\url{https://hlfs.hessen.de/sites/hlfs.hessen.de/files/2022-09/FwDV%20100%20-%20Homepage.pdf}},
  urldate      = {2025-11-14}
}

@INPROCEEDINGS{surmann2023,
    booktitle={2023 IEEE International Symposium on Safety, Security, and Rescue Robotics (SSRR)}, 
    author = {Surmann, Hartmut and Digakis, Niklas and Kremer, Jan-Nicklas and Meine, Julien and Schulte, Max and Voigt, Niklas},
    year = {2023},
    month = {Nov},
    pages = {13-18},
    title = {Redefining Recon: Bridging Gaps with UAVs, 360° Cameras, and Neural Radiance Fields},
    url       = {https://doi.org/10.1109/SSRR59696.2023.10499936}
}

@ARTICLE{nisa2025,
  title     = "A decade of research in small object detection: a comprehensive
               bibliometric analysis",
  author    = "Nisa, Ume and Pozi, Muhammad Syafiq Mohd and Saip, Mohamed Ali",
  journal   = "Int. J. Data Sci. Anal.",
  publisher = "Springer Science and Business Media LLC",
  volume    =  20,
  number    =  8,
  pages     = "7331--7355",
  month     =  dec,
  year      =  2025,
  copyright = "https://creativecommons.org/licenses/by/4.0",
  language  = "en",
  url       = "https://doi.org/10.1007/s41060-025-00881-1"
}

@article{brack2025,
author = {Brack, Ismael V. and Ferrara, Camila and Forero-Medina, German and Domic-Rivadeneira, Enrique and Torrico, Omar and Wanovich, Kevin Thomas and Wilkinson, Benjamin and Valle, Denis},
title = {Counting animals in orthomosaics from aerial imagery: Challenges and future directions},
journal = {Methods in Ecology and Evolution},
volume = {16},
number = {6},
pages = {1051-1060},
doi = {https://doi.org/10.1111/2041-210X.70043},
url = {https://besjournals.onlinelibrary.wiley.com/doi/abs/10.1111/2041-210X.70043},
eprint = {https://besjournals.onlinelibrary.wiley.com/doi/pdf/10.1111/2041-210X.70043},
year = {2025}
}

@article{fonod2025,
   title={Advanced computer vision for extracting georeferenced vehicle trajectories from drone imagery},
   volume={178},
   ISSN={0968-090X},
   url={http://dx.doi.org/10.1016/j.trc.2025.105205},
   DOI={10.1016/j.trc.2025.105205},
   journal={Transportation Research Part C: Emerging Technologies},
   publisher={Elsevier BV},
   author={Fonod, Robert and Cho, Haechan and Yeo, Hwasoo and Geroliminis, Nikolas},
   year={2025},
   month=sep, pages={105205} 
}

@inproceedings{lin2025,
 author = {Lin, Haotong and Chen, Sili and Liew, Jun Hao and Chen, Donny Y. and Li, Zhenyu and Zhao, Yang and Peng, Sida and Guo, Hengkai and Zhou, Xiaowei and Shi, Guang and Feng, Jiashi and Kang, Bingyi},
 booktitle = {International Conference on Learning Representations},
 editor = {C. Vondrick and B. Hariharan and C. Raffel and L. Pinto and D. Yang and A. Faust},
 pages = {141261--141285},
 title = {Depth Anything 3: Recovering the Visual Space from Any Views},
 volume = {2026},
 year = {2026}
}

@article{zhu2021,
  title={Detection and tracking meet drones challenge},
  author={Zhu, Pengfei and Wen, Longyin and Du, Dawei and Bian, Xiao and Fan, Heng and Hu, Qinghua and Ling, Haibin},
  journal={IEEE Transactions on Pattern Analysis and Machine Intelligence},
  volume={44},
  number={11},
  pages={7380--7399},
  year={2021},
  publisher={IEEE}
}

@techreport{bbk2024,
  author      = {{BBK}},
  title       = {{Empfehlungen für Gemeinsame Regelungen zum Einsatz von 
                 Drohnen im Bevölkerungsschutz (EGRED}~2)},
  institution = {Bundesamt für Bevölkerungsschutz und Katastrophenhilfe (BBK)},
  address     = {Bonn},
  year        = {2024},
  note        = {Stand: Juni 2024},
  url         = {https://www.bbk.bund.de/SharedDocs/Downloads/DE/Mediathek/
                 Publikationen/Krisenmanagement/EGRED2.pdf}
}

@article{bi2026,
title = {DRM-YOLO: A YOLOv11-based structural optimization method for small object detection in UAV aerial imagery},
journal = {Image and Vision Computing},
volume = {167},
pages = {105894},
year = {2026},
issn = {0262-8856},
doi = {https://doi.org/10.1016/j.imavis.2025.105894},
url = {https://www.sciencedirect.com/science/article/pii/S0262885625004822},
author = {Hongbo Bi and Rui Dai and Fengyang Han and Cong Zhang},
}

@Article{ning2024,
author={Ning, Tao
and Wu, Wantong
and Zhang, Jin},
title={Small object detection based on YOLOv8 in UAV perspective},
journal={Pattern Analysis and Applications},
year={2024},
month={Aug},
day={18},
volume={27},
number={3},
pages={103},
issn={1433-755X},
doi={10.1007/s10044-024-01323-7},
url={https://doi.org/10.1007/s10044-024-01323-7}
}

@misc{Sapkota2026,
      title={Ultralytics YOLO Evolution: An Overview of YOLO26, YOLO11, YOLOv8 and YOLOv5 Object Detectors for Computer Vision and Pattern Recognition}, 
      author={Ranjan Sapkota and Manoj Karkee},
      year={2026},
      eprint={2510.09653},
      archivePrefix={arXiv},
      primaryClass={cs.CV},
      url={https://arxiv.org/abs/2510.09653}, 
}

@misc{lu2025,
      title={MASF-YOLO: An Improved YOLOv11 Network for Small Object Detection on Drone View}, 
      author={Liugang Lu and Dabin He and Congxiang Liu and Zhixiang Deng},
      year={2025},
      eprint={2504.18136},
      archivePrefix={arXiv},
      primaryClass={cs.CV},
      url={https://arxiv.org/abs/2504.18136}, 
}

@inproceedings{akyon2022,
   title={Slicing Aided Hyper Inference and Fine-Tuning for Small Object Detection},
   url={http://dx.doi.org/10.1109/ICIP46576.2022.9897990},
   DOI={10.1109/icip46576.2022.9897990},
   booktitle={2022 IEEE International Conference on Image Processing (ICIP)},
   publisher={IEEE},
   author={Akyon, Fatih Cagatay and Onur Altinuc, Sinan and Temizel, Alptekin},
   year={2022},
   month=Oct, 
   pages={966--970}
}

@misc{jocher2023,
    author = {Jocher, Glenn and Qiu, Jing and Chaurasia, Ayush},
    license = {AGPL-3.0},
    month = jan,
    title = {{Ultralytics YOLO}},
    url = {https://github.com/ultralytics/ultralytics},
    version = {8.0.0},
    year = {2023}
}

@Article{sun2024,
    author={Sun, Zhaojie
    and Wang, Xuan
    and Zhang, Youlei
    and Song, Yongchao
    and Zhao, Jindong
    and Xu, Jindong
    and Yan, Weiqing
    and Lv, Cuicui},
    title={A comprehensive review of pedestrian re-identification based on deep learning},
    journal={Complex {\&} Intelligent Systems},
    year={2024},
    month={Apr},
    day={01},
    volume={10},
    number={2},
    pages={1733-1768},
    issn={2198-6053},
    doi={10.1007/s40747-023-01229-7},
    url={https://doi.org/10.1007/s40747-023-01229-7}
}

@article{li2025,
    author = {Li, Suya and Ren, Hengyi and Xie, Xin and Cao, Ying},
    title = {A Review of Multi-Object Tracking in Recent Times},
    journal = {IET Computer Vision},
    volume = {19},
    number = {1},
    pages = {e70010},
    doi = {https://doi.org/10.1049/cvi2.70010},
    url = {https://ietresearch.onlinelibrary.wiley.com/doi/abs/10.1049/cvi2.70010},
    eprint = {https://ietresearch.onlinelibrary.wiley.com/doi/pdf/10.1049/cvi2.70010},
    year = {2025}
}

@article{ciaparrone2020,
   title={Deep learning in video multi-object tracking: A survey},
   volume={381},
   ISSN={0925-2312},
   url={http://dx.doi.org/10.1016/j.neucom.2019.11.023},
   DOI={10.1016/j.neucom.2019.11.023},
   journal={Neurocomputing},
   publisher={Elsevier BV},
   author={Ciaparrone, Gioele and Luque Sánchez, Francisco and Tabik, Siham and Troiano, Luigi and Tagliaferri, Roberto and Herrera, Francisco},
   year={2020},
   month=Mar, pages={61--88} 
}

@inproceedings{kostiv2025,
   title={Self-Supervised Real-Time Tracking of Military Vehicles in Low-FPS UAV Footage},
   url={http://dx.doi.org/10.1109/ICMCIS64378.2025.11047873},
   DOI={10.1109/icmcis64378.2025.11047873},
   booktitle={2025 International Conference on Military Communication and Information Systems (ICMCIS)},
   publisher={IEEE},
   author={Kostiv, Markiyan and Adamovskyi, Anatolii and Cherniavskyi, Yevhen and Varenyk, Mykyta and Viniavskyi, Ostap and Krashenyi, Igor and Dobosevych, Oles},
   year={2025},
   month=May, pages={1--8} 
}

@misc{aharon2022,
      title={BoT-SORT: Robust Associations Multi-Pedestrian Tracking}, 
      author={Nir Aharon and Roy Orfaig and Ben-Zion Bobrovsky},
      year={2022},
      eprint={2206.14651},
      archivePrefix={arXiv},
      primaryClass={cs.CV},
      url={https://arxiv.org/abs/2206.14651}, 
}

@misc{ji2026,
      title={View-Centric Multi-Object Tracking with Homographic Matching in Moving UAV}, 
      author={Deyi Ji and Lanyun Zhu and Siqi Gao and Qi Zhu and Yiru Zhao and Peng Xu and Yue Ding and Hongtao Lu and Jieping Ye and Feng Wu and Feng Zhao},
      year={2026},
      eprint={2403.10830},
      archivePrefix={arXiv},
      primaryClass={cs.CV},
      url={https://arxiv.org/abs/2403.10830}, 
}

@misc{kiefer2023,
      title={Memory Maps for Video Object Detection and Tracking on UAVs}, 
      author={Benjamin Kiefer and Yitong Quan and Andreas Zell},
      year={2023},
      eprint={2303.03508},
      archivePrefix={arXiv},
      primaryClass={cs.CV},
      url={https://arxiv.org/abs/2303.03508}, 
}

@misc{yang2023,
      title={Sea You Later: Metadata-Guided Long-Term Re-Identification for UAV-Based Multi-Object Tracking}, 
      author={Cheng-Yen Yang and Hsiang-Wei Huang and Zhongyu Jiang and Heng-Cheng Kuo and Jie Mei and Chung-I Huang and Jenq-Neng Hwang},
      year={2023},
      eprint={2311.03561},
      archivePrefix={arXiv},
      primaryClass={cs.CV},
      url={https://arxiv.org/abs/2311.03561}, 
}

@misc{wu2025,
      title={An Evaluation of DUSt3R/MASt3R/VGGT 3D Reconstruction on Photogrammetric Aerial Blocks}, 
      author={Xinyi Wu and Steven Landgraf and Markus Ulrich and Rongjun Qin},
      year={2025},
      eprint={2507.14798},
      archivePrefix={arXiv},
      primaryClass={cs.CV},
      url={https://arxiv.org/abs/2507.14798}, 
}

@inproceedings{Schoenberger2016,
    author={Sch\"{o}nberger, Johannes Lutz and Frahm, Jan-Michael},
    title={Structure-from-Motion Revisited},
    booktitle={Conference on Computer Vision and Pattern Recognition (CVPR)},
    year={2016},
}

@misc{he2021,
      title={TransReID: Transformer-based Object Re-Identification}, 
      author={Shuting He and Hao Luo and Pichao Wang and Fan Wang and Hao Li and Wei Jiang},
      year={2021},
      eprint={2102.04378},
      archivePrefix={arXiv},
      primaryClass={cs.CV},
      url={https://arxiv.org/abs/2102.04378}, 
}

@misc{oquab2024,
      title={DINOv2: Learning Robust Visual Features without Supervision}, 
      author={Maxime Oquab and Timothée Darcet and Théo Moutakanni and Huy Vo and Marc Szafraniec and Vasil Khalidov and Pierre Fernandez and Daniel Haziza and Francisco Massa and Alaaeldin El-Nouby and Mahmoud Assran and Nicolas Ballas and Wojciech Galuba and Russell Howes and Po-Yao Huang and Shang-Wen Li and Ishan Misra and Michael Rabbat and Vasu Sharma and Gabriel Synnaeve and Hu Xu and Hervé Jegou and Julien Mairal and Patrick Labatut and Armand Joulin and Piotr Bojanowski},
      year={2024},
      eprint={2304.07193},
      archivePrefix={arXiv},
      primaryClass={cs.CV},
      url={https://arxiv.org/abs/2304.07193}, 
}

@ARTICLE{lee2025,
  author={Lee, Min Young and Lee, Christina Dao Wen and Li, Jianghao and Ang, Marcelo H.},
  journal={IEEE Robotics and Automation Letters}, 
  title={DINO-MOT: 3D Multi-Object Tracking With Visual Foundation Model for Pedestrian Re-Identification Using Visual Memory Mechanism}, 
  year={2025},
  volume={10},
  number={2},
  pages={1202-1208},
  doi={10.1109/LRA.2024.3500882}
}

@misc{he2025dino,
      title={DINO-CoDT: Multi-class Collaborative Detection and Tracking with Vision Foundation Models}, 
      author={Xunjie He and Christina Dao Wen Lee and Meiling Wang and Chengran Yuan and Zefan Huang and Yufeng Yue and Ang, Jr., Marcelo H.},
      year={2025},
      eprint={2506.07375},
      archivePrefix={arXiv},
      primaryClass={cs.CV},
      url={https://arxiv.org/abs/2506.07375}, 
}

@misc{zhao2024,
      title={Instance Tracking in 3D Scenes from Egocentric Videos}, 
      author={Yunhan Zhao and Haoyu Ma and Shu Kong and Charless Fowlkes},
      year={2024},
      eprint={2312.04117},
      archivePrefix={arXiv},
      primaryClass={cs.CV},
      url={https://arxiv.org/abs/2312.04117}, 
}

@misc{huang2025,
      title={Generalizable Object Re-Identification via Visual In-Context Prompting}, 
      author={Zhizhong Huang and Xiaoming Liu},
      year={2025},
      eprint={2508.21222},
      archivePrefix={arXiv},
      primaryClass={cs.CV},
      url={https://arxiv.org/abs/2508.21222}, 
}

@misc{simeoni2025,
      title={DINOv3}, 
      author={Oriane Siméoni and Huy V. Vo and Maximilian Seitzer and Federico Baldassarre and Maxime Oquab and Cijo Jose and Vasil Khalidov and Marc Szafraniec and Seungeun Yi and Michaël Ramamonjisoa and Francisco Massa and Daniel Haziza and Luca Wehrstedt and Jianyuan Wang and Timothée Darcet and Théo Moutakanni and Leonel Sentana and Claire Roberts and Andrea Vedaldi and Jamie Tolan and John Brandt and Camille Couprie and Julien Mairal and Hervé Jégou and Patrick Labatut and Piotr Bojanowski},
      year={2025},
      eprint={2508.10104},
      archivePrefix={arXiv},
      primaryClass={cs.CV},
      url={https://arxiv.org/abs/2508.10104}, 
}

\end{document}